\documentclass[nohyperref]{article}

\usepackage{microtype}
\usepackage{graphicx}
\usepackage{subfigure}
\usepackage{booktabs} 

\usepackage{graphicx}
\usepackage{amsmath}
\usepackage{amssymb}
\usepackage{booktabs}

\usepackage{amsfonts}       
\usepackage{nicefrac}       
\usepackage{microtype}      
\usepackage{xcolor}         
\usepackage{microtype}      
\usepackage{graphicx}
\usepackage{subfigure}
\usepackage{amsmath,amssymb}
\usepackage{mathrsfs}
\usepackage{multirow}
\usepackage{float}
\usepackage{setspace}
\usepackage{mathrsfs} 
\usepackage{enumitem}

\usepackage{hyperref}

\usepackage[accepted]{icml2022}

\usepackage{amsmath}
\usepackage{amssymb}
\usepackage{mathtools}
\usepackage{amsthm}

\usepackage[capitalize,noabbrev]{cleveref}

\theoremstyle{plain}

\theoremstyle{definition}

\theoremstyle{remark}

\usepackage[textsize=tiny]{todonotes}

\icmltitlerunning{NP-Match: When Neural Processes meet Semi-Supervised Learning}

\begin{document}

\twocolumn[
\icmltitle{NP-Match: When Neural Processes meet Semi-Supervised Learning}



\icmlsetsymbol{equal}{*}

\begin{icmlauthorlist}
\icmlauthor{Jianfeng Wang}{yyy}
\icmlauthor{Thomas Lukasiewicz}{tuw,yyy}
\icmlauthor{Daniela Massiceti}{comp1}
\icmlauthor{Xiaolin Hu}{sch}
\icmlauthor{Vladimir Pavlovic}{sch1} 
\icmlauthor{Alexandros Neophytou}{comp2} 
\end{icmlauthorlist}

\icmlaffiliation{yyy}{Department of Computer Science, University of Oxford, UK.}
\icmlaffiliation{tuw}{Institute of Logic and Computation, TU Wien, Austria.}
\icmlaffiliation{comp1}{Microsoft Research, Cambridge, UK.}
\icmlaffiliation{sch}{Department of Computer Science and Technology, Tsinghua University, Beijing, China.}
\icmlaffiliation{sch1}{Department of Computer Science, Rutgers University, New Jersey, USA.}
\icmlaffiliation{comp2}{Microsoft, Applied Science Group, Reading, UK}

\icmlcorrespondingauthor{Jianfeng Wang}{jianfeng.wang@cs.ox.ac.uk}

\icmlkeywords{Machine Learning, ICML}

\vskip 0.3in
]



\printAffiliationsAndNotice{} 

\begin{abstract}
Semi-supervised learning (SSL) has been widely explored in recent years, and it is an effective way of leveraging unlabeled data to reduce the reliance on labeled data. In this work, we adjust neural processes (NPs) to the  semi-supervised image classification task, resulting in a new method named NP-Match. NP-Match is suited to this task for two reasons. 
Firstly, NP-Match implicitly compares data points when making predictions, and as a result, the prediction of each unlabeled data point is affected by the labeled data points that are similar to it, which improves the quality of pseudo-labels. 
Secondly, NP-Match is able to estimate uncertainty that can be used as a tool for selecting unlabeled samples with reliable pseudo-labels. Compared with uncertainty-based SSL methods implemented with Monte Carlo (MC) dropout, NP-Match estimates uncertainty with much less computational overhead, which can save time at both the training and the testing phases. We conducted extensive experiments on four public datasets, and NP-Match outperforms state-of-the-art (SOTA) results or achieves competitive results on them, which shows the effectiveness of NP-Match and its potential for SSL. 
\end{abstract} 
\section{Introduction}
\label{intro}
Deep neural networks have been widely used in computer vision tasks \cite{krizhevsky2012imagenet, simonyan2014very, szegedy2015going, szegedy2016rethinking, he2016deep} due to their strong performance. Training deep neural networks relies on large-scale labeled datasets, but annotating large-scale datasets is time-consuming, which encourages researchers to explore semi-supervised learning (SSL). SSL aims to learn from few labeled data and a large amount of unlabeled data, and it has been a long-standing problem in computer vision and machine learning \cite{sohn2020fixmatch, zhang2021flexmatch, rizve2021defense, pham2021meta, li2014towards, liu2010large, berthelot2019mixmatch, berthelot2019remixmatch}. In this work, we focus on SSL for image classification. 

Most recent approaches to SSL for image classification are based on the combination of consistency regularization and pseudo-labeling \cite{sohn2020fixmatch, li2021comatch, rizve2021defense, zhang2021flexmatch, nassar2021all, pham2021meta, hu2021simple}. They  can be further classified into two categories, namely,  deterministic \cite{sohn2020fixmatch, li2021comatch, zhang2021flexmatch, nassar2021all, pham2021meta, hu2021simple} and probabilistic ones \cite{rizve2021defense}. 
A deterministic approach aims at directly making predictions, while a probabilistic approach tries to additionally model the predictive distribution, such as using Bayesian neural networks (BNNs), which are implemented by Monte Carlo (MC) dropout \cite{gal2016dropout}. 
As a result, the former cannot estimate the uncertainty of the model's prediction, and unlabeled samples are selected only based on  high-confidence predictions. In contrast, the latter can give uncertainties for unlabeled samples, and the uncertainties can be combined with high-confidence predictions for picking or refining pseudo-labels.

Current SOTA methods for the semi-supervised image classification task are deterministic, including FixMatch \cite{sohn2020fixmatch}, CoMatch \cite{li2021comatch}, and FlexMatch \cite{zhang2021flexmatch}, which have achieved promising results on public benchmarks. 
In contrast, progress on probabilistic approaches 
lags behind, which is mainly shown by the fact that there are only few studies on this task and MC dropout becomes the only option for implementing the probabilistic model \cite{rizve2021defense}. In addition, 
MC dropout also dominates the uncertainty-based approaches to other SSL tasks \cite{sedai2019uncertainty, shi2021inconsistency, wang2021tripled, yu2019uncertainty, zhu2020grasping}. MC dropout, however, is time-consuming, requiring several feedforward passes to get uncertainty at both the training and the testing stages, especially when some large models are used. 

To solve this drawback and to further promote the related research, we need to find better probabilistic approaches for SSL. 
Considering that MC dropout is an approximation to the Gaussian process (GP) model \cite{gal2016dropout}, we turn to another approximation model called
neural processes (NPs) \cite{garnelo2018neural}, which can be regarded as an NN-based formulation that approximates GPs.
Similarly to a GP, a neural process  is also a probabilistic model that defines distributions over functions.
Thus, an NP is able to rapidly adapt to new observations, with the advantage of estimating the uncertainty of each observation.
There are two main aspects that motivate us to investigate NPs in SSL. 
Firstly, GPs have been preliminarily explored for some SSL tasks \cite{sindhwani2007semi, jean2018semi, yasarla2020syn2real}, because of the property that their kernels are able to compare labeled data with unlabeled data when making predictions. 
NPs share this property, since it has been proved that NPs can learn non-trivial implicit kernels from data \cite{garnelo2018neural}. As a result, NPs are able to make predictions for target points conditioned on context points. This feature is highly relevant to SSL, which must learn from limited labeled samples in order to make predictions for unlabeled data, similarly to how NPs are able to impute unknown pixel values (i.e., target points) when given only a small number of known pixels (namely, context points) \cite{garnelo2018neural}. 
Due to the learned implicit kernels in NPs \cite{garnelo2018neural} and the successful application of GPs to different SSL tasks \cite{sindhwani2007semi, jean2018semi, yasarla2020syn2real}, NPs could be a suitable probabilistic model for SSL, as the kernels can compare labeled data with unlabeled data in order to improve the quality of pseudo-labels for the unlabeled data at the training stage. Secondly, previous GP-based works for SSL do not explore the semi-supervised large-scale image classification task, since GPs are computationally expensive, which usually incur a $\mathcal{O}(n^3)$ runtime for $n$ training points. But, unlike GPs, NPs are more efficient than GPs, providing the possibility of applying NPs to this task. NPs are also computationally significantly more efficient than current MC-dropout-based approaches to SSL, since, given an input image, they only need to perform one feedforward pass to obtain the prediction with an uncertainty estimate.

In this work, we take the first step to explore NPs in large-scale semi-supervised image classification, and propose a new probabilistic method called NP-Match. NP-Match still rests on the combination of consistency regularization and pseudo-labeling, but it incorporates NPs to the top of deep neural networks, and therefore it is a probabilistic approach. Compared to the previous probabilistic method for semi-supervised image classification \cite{rizve2021defense}, NP-Match not only can make predictions and estimate uncertainty more efficiently,  inheriting the advantages of NPs, but also can achieve a better performance on public benchmarks. 

Summarizing, the main contributions of this paper are:\vspace*{-1ex}
\begin{itemize}[leftmargin=*, itemsep=0.5pt]
\item We propose NP-Match, which adjusts NPs to SSL, and explore its use in semi-supervised large-scale image classification. To our knowledge, this is the first such work. In addition, NP-Match has the potential to break the monopoly of MC dropout as the probabilistic model in SSL. 

\item We experimentally show that the Kullback-Leibler (KL) divergence in the evidence lower bound (ELBO) of NPs \cite{garnelo2018neural} is not a good choice in the context of SSL, which may negatively impact the learning of global latent variables. To tackle this problem, we propose a new uncertainty-guided skew-geometric Jensen-Shannon (JS) divergence ($JS^{G_{\alpha_u}}$) for NP-Match. 

\item We show that NP-Match outperforms SOTA results or achieves competitive results on four public benchmarks, demonstrating its effectiveness for SSL. We also  show that NP-Match estimates uncertainty faster than the MC-dropout-based probabilistic model, which can improve the training and the test efficiency.  
\end{itemize}

The rest of this paper is organized as follows. In Section~\ref{sec:related_work}, we review related methods. Section~\ref{sec:methodology} presents NP-Match and the uncertainty-guided skew-geometric JS divergence ($JS^{G_{\alpha_u}}$),  followed by the experimental settings and results in Section~\ref{sec:experiment}. In Section~\ref{sec:conclusion}, we give a summary and an outlook on future research. The source code is available at: \href{https://github.com/Jianf-Wang/NP-Match}{https://github.com/Jianf-Wang/NP-Match}.

\section{Related Work}
\label{sec:related_work} 

We now briefly review related works, including {\it semi-super\-vised learning (SSL) for image classification}, {\it Gaussian processes (GPs) for SSL}, and {\it neural processes (NPs)}.

{\bf SSL for image classification.} 
Most methods for semi-supervised image classification in the past few years are based on pseudo-labeling and consistency regularization. Pseudo-labeling approaches
rely on the high confidence of pseudo-labels, which can be added to the training data set as labeled data, and those approaches can be classified into two classes, namely, disagreement-based models and self-training models. The former models aim to train multiple learners and exploit the disagreement during the learning process \cite{qiao2018deep, dong2018tri}, while the latter models aim at training the model on a small amount of labeled data, and then using its predictions on the unlabeled data as pseudo-labels \cite{lee2013pseudo, zhai2019s4l, wang2020enaet, pham2021meta}. 
Consistency-regularization-based approaches work by performing different transformations on an input image and adding a regularization term to make their predictions consistent \cite{bachman2014learning, sajjadi2016regularization, laine2016temporal, berthelot2019mixmatch, xie2019unsupervised}. 
Based on these two approaches, FixMatch \cite{sohn2020fixmatch} is proposed, which achieves new state-of-the-art (SOTA) results on the most commonly-studied SSL benchmarks.
FixMatch \cite{sohn2020fixmatch} combines the merits of these two approaches: given an unlabeled image, weak data augmentation and strong data augmentation are performed on the image, leading to two versions of the image, and then FixMatch produces a pseudo-label based on its weakly-augmented version and a preset confidence threshold, which is used as the true label for its strongly augmented version to train the whole framework.  
The success of FixMatch inspired several subsequent methods \cite{li2021comatch, rizve2021defense, zhang2021flexmatch, nassar2021all, pham2021meta, hu2021simple}. For instance, \citet{li2021comatch} additionally design the classification head and the projection head for generating a class probability and a low-dimensional embedding, respectively. The projection head and the classification head are jointly optimized during training. 
Specifically, the former is learnt with contrastive learning on pseudo-label graphs to encourage the embeddings of samples with similar pseudo-labels to be close, and the latter is trained with pseudo-labels that are smoothed by aggregating information from nearby samples in the embedding space. \citet{zhang2021flexmatch} propose to use dynamic confidence thresholds that are automatically adjusted according to the model’s learning status of 
each~class. \citet{rizve2021defense} propose an uncertainty-aware
pseudo-label selection (UPS) framework for semi-supervised image classification. The UPS framework introduces MC dropout to obtain uncertainty estimates, which are then leveraged as a tool for selecting pseudo-labels. This is the first work using MC dropout for semi-supervised image classification. 

{\bf GPs for SSL.}  Since NPs are also closely related to GPs, we review the application of GPs to different SSL tasks in this part.
GPs, which are non-parametric models, have been preliminarily investigated in different semi-supervised learning tasks. For example, \citet{sindhwani2007semi} introduce a semi-supervised GP classifier, which incorporates the information of relationships among labeled and unlabeled data  into the kernel. Their approach, however, has high  computational costs and is thus only evaluated for a simple binary classification task on small datasets.  Deep kernel learning  \cite{wilson2016deep} also lies on the spectrum between NNs and GPs, and has been integrated into a new framework for the semi-supervised regression task, named semi-supervised deep kernel learning  \cite{jean2018semi}, which  aims to minimize the predictive variance for unlabeled data,  encouraging  unlabeled embeddings to be near labeled embeddings. Semi-supervised deep kernel learning, however, has not been applied to SSL image classification, and (similarly to semi-supervised GPs) also comes with a high (cubic) computational complexity. 
Recently, \citet{yasarla2020syn2real} proposed to combine GPs with UNet \cite{ronneberger2015u} for SSL image deraining. Here, GPs are used to get pseudo-labels for unlabeled samples based on the feature representations of labeled and unlabeled images. GPs have also been combined with graph convolutional networks for  semi-supervised learning on graphs \cite{ng2018bayesian, walker2019graph, liu2020uncertainty}. 
Although many previous works explore GPs in different semi-supervised learning tasks, none of them investigates the application of GPs to  semi-supervised large-scale image classification.

{\bf NPs.} The origin of NPs can be traced back to conditional NPs  \cite{garnelo2018conditional}, which define conditional distributions over functions given a set of observations. Conditional NPs, however, do not introduce global latent variables for observations, which led to the birth of NPs \cite{garnelo2018neural}. 
In NPs, the mean-aggregator is used to summarize the encoded inputs of a task into a global latent variable, which is then used to make predictions on targets in the task. In recent years, several NP variants have been proposed to better approximate stochastic processes. 
For example, \citet{kim2019attentive} consider that the mean-aggregator may cause difficulties for the decoder to pick relevant information with regard to making predictions, and they introduce a differentiable attention mechanism to solve this issue, resulting in new attentive NPs. 
\citet{gordon2019convolutional} consider that the translation equivariance is important for prediction problems, which should be considered. Therefore, they incorporate translation equivariance into NPs and design a new model,  called convolutional conditional NPs. 
Besides, \citet{louizos2019functional} consider that using global latent variables is not flexible for encoding inductive biases. Thus, they propose to use local latent variables along with a dependency structure among them, resulting in new functional NPs. 
\citet{lee2020bootstrapping} point out the limitation of using a single Gaussian latent variable to model functional uncertainty. To solve the limitation, they propose to use bootstrapping for inducing functional uncertainty, leading to a new NP variant, called Bootstrapping Neural Processes (BNP). 
\citet{bruinsma2021gaussian} introduce a novel member of the NP family that incorporates translation
equivariance and models the predictive distributions directly with Gaussian processes, called Gaussian NP (GNP). GNPs do not allow for correlations in the predictive distribution, but also provide universal approximation guarantees. 
Currently, NPs and their variants have been widely used in many different settings, including meta-learning \cite{singh2019sequential, yoon2020robustifying, requeima2019fast} and sequential data modelling \cite{qin2019recurrent}, but they have not been studied in SSL. Our work is the first to leverage NPs for semi-supervised large-scale image recognition. We choose the most basic model from \citep{garnelo2018neural} (i.e., the original NPs), and we expect that future works can further study the application of other variants to this task. 

\begin{figure*}[t]
\centering
\includegraphics[width=0.95\linewidth]{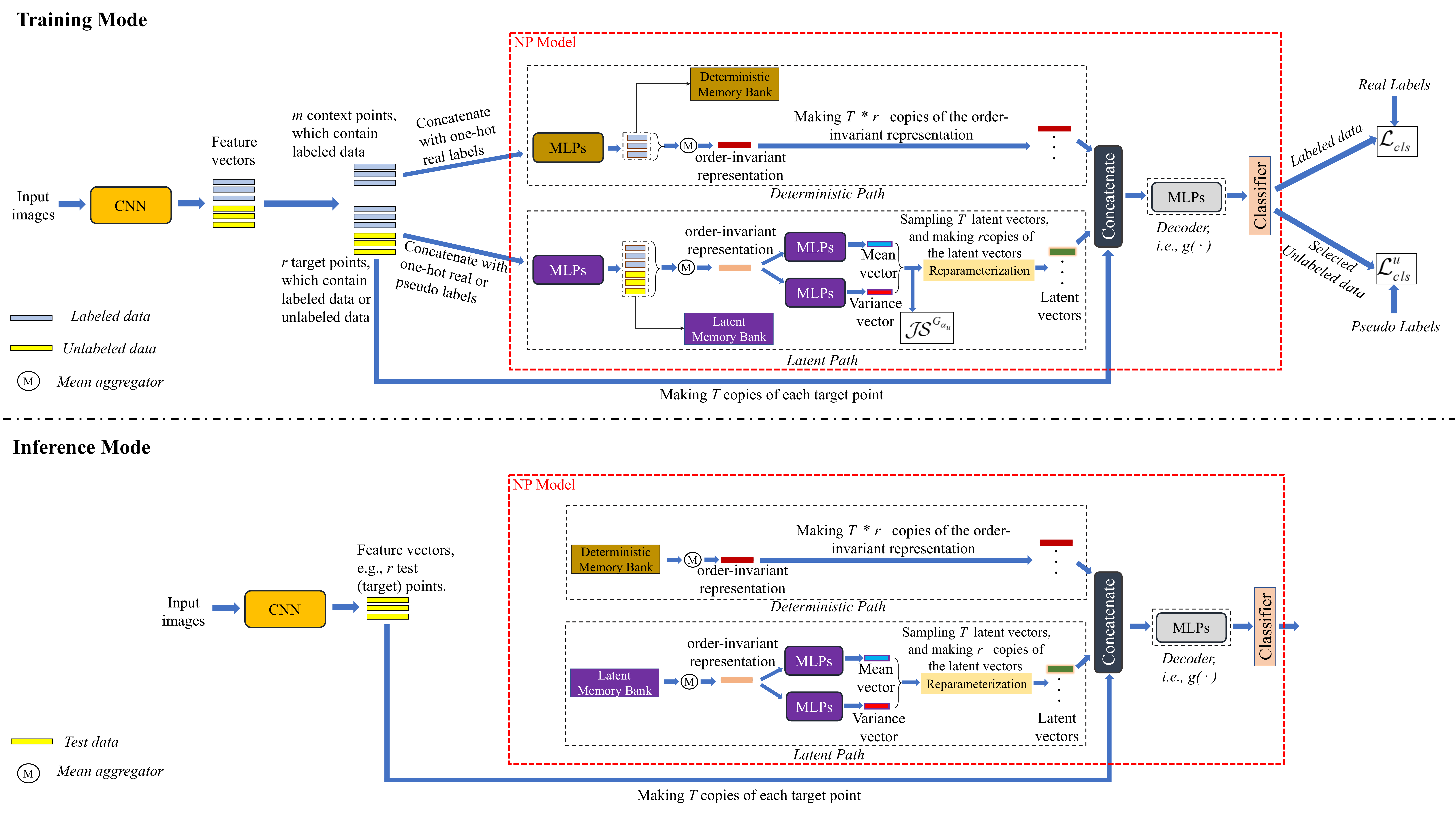}
\vspace{-1ex}
\caption{
Overview of NP-Match: it contains a convolutional neural network (CNN) and an NP model that is shown in the red dotted box. The feature vectors come from the global average pooling layer in the CNN.}
\label{fig:np} 
\end{figure*}

\section{Methodology}
\label{sec:methodology}  
In this section, we provide a brief introduction to neural processes (NPs) and a detailed description of our NP-Match.  

\subsection{NPs}

NPs approximate stochastic processes via finite-dimensional marginal distributions \cite{garnelo2018neural}. Formally, given a probability space $(\Omega, \Sigma, \Pi)$ and an index set $\mathcal{X}$, a stochastic process can be  written as $\{F(x, \omega) : x \in \mathcal{X} \}$, where $F(\cdot\ , \ \omega)$ is a sample function
mapping  $\mathcal{X}$ to another space $\mathcal{Y}$ for any point $\omega \in \Omega$. For each finite sequence $x_{1:n}$, a marginal joint distribution function can be defined on the function values $F(x_1, \ \omega), F(x_2, \ \omega), \ldots, F(x_n, \ \omega)$, which satisfies two conditions given by the Kolmogorov Extension Theorem \cite{oksendal2003stochastic}, namely, exchangeability and consistency. 
Assuming that a density $\pi$, where $d\Pi = \pi d\mu$, and the likelihood density $p(y_{1:n}|F(\cdot\ , \ \omega), x_{1:n})$ exist, the marginal joint distribution function can be written~as:
\begin{equation}
\small
\label{eq:joint1}
p(y_{1:n}|x_{1:n}) = \int \pi(\omega) p(y_{1:n}|F(\cdot\ , \ \omega), x_{1:n}) d\mu(\omega).
\end{equation}
The exchangeability condition requires the joint distributions to be invariant to permutations of the elements, i.e., $p(y_{1:n}|x_{1:n}) = p(\varphi(y_{1:n})|\varphi(y_{1:n}))$, where $\varphi$ is a permutation of $\{1, \ldots,n\}$. 
The consistency condition expresses  that if a part of the sequence is marginalised out, then the resulting marginal distribution is consistent with that defined on the original sequence. 

Letting $(\Omega, \Sigma)$ be $(\mathbb{R}^d, \mathcal{B}(\mathbb{R}^d))$,  where $\mathcal{B}(\mathbb{R}^d)$ denotes the {\it Borel} $\sigma${\it -algebra} of $\mathbb{R}^d$, NPs parameterize the function $F(\cdot\ , \ \omega)$ with a high-dimensional random vector $z$ sampled from a multivariate Gaussian distribution. Then, $F(x_i , \ \omega)$ can be replaced by $g(x_i, z)$, where $g(\cdot)$ denotes a neural network, and Eq.~(\ref{eq:joint1}) becomes:
\begin{equation}
\small
\label{eq:joint2}
p(y_{1:n}|x_{1:n}) = \int \pi(z)p(y_{1:n}|g(x_{1:n}, z), x_{1:n}) d\mu(z).
\end{equation} 
The training objective of NPs is to maximize $p(y_{1:n}|x_{1:n})$, and the learning procedure reflects the NPs' property that they have the capability to make predictions for target points conditioned on context points \cite{garnelo2018neural}. 

\subsection{NP-Match}
As Figure~\ref{fig:np} shows, NP-Match is mainly composed of two parts: a deep neural network and an NP model. The deep neural network is leveraged for obtaining feature representations of input images, while the NP model is built upon the network to receive the representations for classification. 

\subsubsection{NP Model for Semi-supervised Image Classification} 
Since we extend the original NPs \cite{garnelo2018neural} to the classification task, $p(y_{1:n}|g(x_{1:n}, z), x_{1:n})$ in Eq.~(\ref{eq:joint2})  should define a categorical distribution rather than a Gaussian distribution. Therefore, we parameterize the categorical distribution by probability vectors from a classifier that contains a weight matrix ($\mathcal{W}$) and a softmax function ($\Phi$):
\begin{equation}
\label{eq:likelihood}
p(y_{1:n}|g(x_{1:n}, z), x_{1:n}) =  Categorical(\Phi(\mathcal{W}g(x_{1:n}, z))).
\end{equation}
Note that $g(\cdot)$ can be learned via amortised variational inference, and to use this method, two steps need to be done: (1)~parameterize a variational distribution over $z$, and  (2)~find the evidence lower bound (ELBO) as the learning objective. For the first step, we let $q(z|x_{1:n}, y_{1:n})$ be a variational distribution defined on the same measure space, which can be parameterized by a neural network. For the second step, 
given a finite sequence with length $n$, we assume that there are $m$ context points ($x_{1:m}$) and $r$ target points ($x_{m+1:\ m+r}$) in it, i.e., $m+r=n$. Then, the ELBO is given by  (with proof in the appendix):
\begin{equation}
\small
\begin{aligned}
\label{eq:elbo}
&log\ p(y_{1:n}|x_{1:n}) \ge \\ 
& \mathbb{E}_{q(z|x_{m+1:\ m+r}, y_{m+1:\ m+r})}\Big[\sum^{m+r}_{i=m+1}log\ p(y_i|z, x_i) - \\
&log\ \frac{q(z|x_{m+1:\ m+r}, y_{m+1:\ m+r})}{q(z|x_{1:m}, y_{1:m})}\Big] + const.
\end{aligned}
\end{equation}
To learn the NP model, one can maximize this ELBO. Under the setting of SSL, we consider that only labeled data can be treated as context points, and either labeled or unlabeled data can be treated as target points, since the target points are what the NP model makes predictions for.

\subsubsection{NP-Match Pipeline} 
 
We now introduce the NP-Match pipeline.
We first focus on the configuration of the NP model, which is shown in the red dotted box in Figure~\ref{fig:np}. 
The NP model is mainly constructed by MLPs, memory banks, and a classifier. Specifically, the classifier is composed of the weight matrix ($\mathcal{W}$) and the softmax function ($\Phi$). Similarly to the original implementation of NPs, we build two paths with the memory banks and MLPs, namely, the latent path and the deterministic path. The decoder $g(\cdot)$ is also implemented with MLPs. 
The workflow of NP-Match at the training stage and the inference stage are different, which are shown in Figure~\ref{fig:np}, and they are introduced separately as follows.

{\bf Training mode.} 
Given a batch of $B$ labeled images  $\mathcal{L} = \{(x_i, y_i): i\,{\in}\, \{1,\ldots,B\}\}$ and a batch of unlabeled images $\mathcal{U} = \{x^u_i: i\,{\in}\, \{1,\ldots, \mu B\}\}$ at each iteration, where $\mu$ determines the relative size of $\mathcal{U}$ to $\mathcal{L}$, we apply weak augmentation (i.e., 
crop-and-flip) on the labeled and unlabeled samples, and strong augmentation (i.e.,  RandAugment \cite{cubuk2020randaugment}) on only the unlabeled samples. 
After the augmentation is applied, the images are passed through the deep neural network, 
and the features are input to the NP model, which finally outputs the predictions and associated uncertainties. The detailed process can be summarized as follows. 
At the start of each iteration, NP-Match is switched to inference mode, and it makes predictions for the weakly-augmented unlabeled data. 
Then, inference mode is turned off, and those predictions are treated as pseudo-labels for unlabeled data. 
After receiving the features, real labels, and pseudo-labels, the NP model first duplicates the labeled samples and treats them as context points, and all the labeled and unlabeled samples in the original batches are then treated as target points, since the NP model needs to make a prediction for them. Thereafter, the target points and context points are separately fed to the latent path and the deterministic path. As for the latent path, target points are concatenated with their corresponding real labels or pseudo labels, and processed by MLPs to get new representations. 
Then, the representations are averaged by a mean aggregator along the batch dimension, leading to an order-invariant representation, which implements the exchangeability  and the consistency condition, and they are simultaneously stored in the latent memory bank, which is updated with a first-in-first-out strategy. 
After the mean aggregator, the order-invariant representation is further processed by other two MLPs in order to get the mean vector and the variance vector, which are used for sampling latent vectors via the reparameterization trick, and the number of latent vectors sampled at each feed-forward pass is denoted $T$. 
As for the deterministic path, context points are input to this path and are processed in the same way as the target points, until an order-invariant representation is procured from the mean aggregator. We also introduce a memory bank to the deterministic path for storing representations.
Subsequently, each target point is concatenated with the $T$ latent vectors and the order-invariant
representations from the deterministic path (note that, practically, the target point and the order-invariant representations from the deterministic path must be copied $T$ times). After the concatenation operation, the $T * r$ feature representations are fed into the decoder $g(\cdot)$ and then the classifier, which outputs $T$ probability distributions over classes for each target point. The final prediction for each target point can be obtained by averaging the $T$ predictions, and the uncertainty is computed as the entropy of the average prediction \cite{kendall2017uncertainties}. 
The ELBO (Eq.~(\ref{eq:elbo})) shows the learning objective.  Specifically, the first term can be achieved by using the cross-entropy loss on the labeled and unlabeled data with their corresponding real labels and pseudo-labels, while the second term is the KL divergence between $q(z|x_{m+1:\ m+r}, y_{m+1:\ m+r})$ and $q(z|x_{1:m}, y_{1:m})$.

{\bf Inference mode.} Concerning a set of test images, they are also passed through the deep neural network at first to obtain their feature representations. Then, they are treated as target points and are fed to the NP model. Since 
the labels of test data are not available, it is impossible to obtain the order-invariant representation from test data. In this case, the stored features in the two memory banks can be directly used. As the bottom diagram of Figure~\ref{fig:np} shows, after the order-invariant representations are obtained from the memory banks, the target points are leveraged in the same way as in the training mode to generate concatenated feature representations for the decoder $g(\cdot)$ and then the~classifier. 

\subsubsection{Uncertainty-guided Skew-Geometric JS Divergence} 
\label{sec:ugjs}
NP-Match, like many SSL approaches, relies on the use of pseudo-labels for the unlabeled samples. Pseudo-labels, however, are sometimes inaccurate and can lead to the neural network learning poor feature representations. In our pipeline, this can go on to impact the representation procured from the mean-aggregator and hence the model's estimated mean vector, variance vector, and global latent vectors (see ``Latent Path" in \cref{fig:np}).
To remedy this, similarly to how the KL divergence term in the ELBO (Eq.~(\ref{eq:elbo})) is used to learn global latent variables \cite{garnelo2018neural}, we propose a new distribution divergence, called the uncertainty-guided skew-geometric JS divergence ($JS^{G_{\alpha_u}}$). We first formalize the definition of $JS^{G_{\alpha_u}}$:
 
{\bf Definition 1.} \emph{Let $(\Omega, \Sigma)$ be a measurable space, where $\Omega$ denotes the sample space, and $\Sigma$ denotes the $\sigma$-algebra of measurable events. $P$ and $Q$ are two probability measures defined on the measurable space. Concerning a positive measure\footnote{Specifically, the positive measure is usually the Lebesgue measure with the Borel $\sigma$-algebra $\mathcal{B}(\mathbb{R}^d)$ or the counting measure with the power set $\sigma$-algebra $2^\Omega$.}, which is denoted as $\mu$, the uncertainty-guided skew-geometric JS divergence ($JS^{G_{\alpha_u}}$) can be defined as: }
\begin{equation}
\small
\begin{aligned}
\label{eq:JS_skew}
&JS^{G_{\alpha_u}}(p, q) = \\
&(1 - \alpha_u)  \int p \ log \frac{p}{G(p,q)_{\alpha_u}} d\mu +  \alpha_u \int q \ log \frac{q}{G(p,q)_{\alpha_u}} d\mu,
\end{aligned}
\end{equation} 
\emph{where $p$ and $q$ are the Radon-Nikodym derivatives of $P$ and $Q$ with respect to $\mu$,  the scalar $\alpha_u \in [0, 1]$ is  calculated based on the uncertainty, and $G(p,q)_{\alpha_u} = p^{1-\alpha_u}q^{\alpha_u}$ $/$ $(\int_{\Omega} p^{1-\alpha_u}q^{\alpha_u} d\mu)$. The dual form of $JS^{G_{\alpha_u}}$ is given by: 
}
\begin{equation}
\small
\begin{aligned}
\label{eq:JS_skew_dual}
&JS_*^{G_{\alpha_u}}(p, q) = (1 - \alpha_u)  \int G(p,q)_{\alpha_u} \ log \frac{G(p,q)_{\alpha_u}}{p} d\mu +  \\
& \alpha_u \int G(p,q)_{\alpha_u} \ log \frac{G(p,q)_{\alpha_u}}{q} d\mu.
\end{aligned}
\end{equation} 

The proposed $JS^{G_{\alpha_u}}$ is an extension of the skew-geo\-met\-ric JS diver\-gence first proposed by \citet{nielsen2020generalization}. Specifically, \citet{nielsen2020generalization} generalizes the JS divergence with abstract means (quasi-arithmetic means \cite{niculescu2006convex}), in which a scalar $\alpha$ is defined to control the degree of divergence skew.\footnote{The divergence skew means how closely related the intermediate distribution (the abstract mean of $p$ and $q$) is to $p$ or $q$.} By selecting the weighted geometric mean $p^{1-\alpha}q^{\alpha}$,  such generalized JS divergence becomes the skew-geometric JS divergence, which can be easily applied to the Gaussian distribution because of its property that the weighted product of exponential family distributions stays in the exponential family \cite{nielsen2009statistical}.  
Our $JS^{G_{\alpha_u}}$ extends such divergence by incorporating the uncertainty into the scalar $\alpha$ to dynamically adjust the divergence skew. We assume the real variational distribution of the global latent variable under the supervised learning to be $q^*$. If the framework is trained with real labels, the condition $q(z|x_{m+1:\ m+r}, y_{m+1:\ m+r})=q(z|x_{1:m}, y_{1:m})=q^*$ will hold after training, since they are all the marginal distributions of the same stochastic process. 
However, as for SSL, $q(z|x_{m+1:\ m+r}, y_{m+1:\ m+r})$ and $q(z|x_{1:m}, y_{1:m})$ are no longer equal to $q^*$, as some low-quality representations are involved during training, which affect the estimation of $q(z|x_{m+1:\ m+r}, y_{m+1:\ m+r})$ and $q(z|x_{1:m}, y_{1:m})$. Our proposed $JS^{G_{\alpha_u}}$ solves this issue by introducing an intermediate distribution that is  calculated via $G(q(z|x_{1:m},y_{1:m}),$ $q(z|x_{m+1:\ m+r},y_{m+1:\ m+r}))_{\alpha_u}$, where $\alpha_u = u_{c_{avg}} / (u_{c_{avg}} + u_{t_{avg}})$.   Here, $u_{c_{avg}}$  denotes the average value over the uncertainties of the predictions of context points, and $u_{t_{avg}}$ represents the average value over that of target points. With this setting, the intermediate distribution is usually close to  $q^*$. For example, when $u_{c_{avg}}$ is large, and $u_{t_{avg}}$ is small, which means that there are many low-quality feature presentations  involved for calculating $q(z|x_{1:m}, y_{1:m})$, and $q(z|x_{m+1:\ m+r}, y_{m+1:\ m+r})$ is closer to $q^*$, then $G(q(z|x_{1:m}, y_{1:m}), q(z|x_{m+1:\ m+r}, y_{m+1:\ m+r}))_{\alpha_u}$ will be close to $q(z|x_{m+1:\ m+r}, y_{m+1:\ m+r})$, and as a result, the network is optimized to learn the distribution of the global latent variable in the direction to $q^*$, which mitigates the issue to some extent.\footnote{As long as one of $q(z|x_{1:m}, y_{1:m})$ and $q(z|x_{m+1:\ m+r},$ $y_{m+1:\ m+r})$ is close to $q*$, the proposed $JS^{G_{\alpha_u}}$ mitigates the issue, but $JS^{G_{\alpha_u}}$ still has difficulties to solve the problem when both of their calculations involve many low-quality representations.}  Concerning  the variational distribution being  supposed to be a Gaussian distribution, we introduce the following theorem (with proof in the appendix) for calculating $JS^{G_{\alpha_u}}$ on Gaussian~distributions:

\smallskip
{\bf Theorem 1.} \emph{Given two multivariate Gaussians $\mathcal{N}_1(\mu_1, \Sigma_1)$ and $\mathcal{N}_2(\mu_2, \Sigma_2)$, the following holds:}  
\begin{equation}
\small
\begin{aligned}
\label{eq:JS_skew_gaussian}
&JS^{G_{\alpha_u}}(\mathcal{N}_1, \mathcal{N}_2) = \frac{1}{2}(tr(\Sigma^{-1}_{\alpha_u}((1 - \alpha_u)\Sigma_1 + \alpha_u\Sigma_2)) + \\
&(1-\alpha_u)(\mu_{\alpha_u} - \mu_1)^T\Sigma^{-1}_{\alpha_u}(\mu_{\alpha_u} - \mu_1) + \\
&\alpha_u(\mu_{\alpha_u} - \mu_2)^T\Sigma^{-1}_{\alpha_u}(\mu_{\alpha_u} - \mu_2) + \\
&log[\frac{det[\Sigma_{\alpha_u}]}{det[\Sigma_1]^{1-\alpha_u} det[\Sigma_2]^{\alpha_u}}] - D)\\
&JS_*^{G_{\alpha_u}}(\mathcal{N}_1, \mathcal{N}_2) = \\
&\frac{1}{2}( log[\frac{det[\Sigma_1]^{1-\alpha_u} det[\Sigma_2]^{\alpha_u}}{det[\Sigma_{\alpha_u}]}] + \alpha_u\mu_2^T\Sigma_2^{-1}\mu_2 \\
 & -\mu_{\alpha_u}^T\Sigma_{\alpha_u}^{-1}\mu_{\alpha_u} + (1 - \alpha_u)\mu_1^T\Sigma_1^{-1}\mu_1),
\end{aligned}
\end{equation}  
\emph{where $\Sigma_{\alpha_u}=((1-\alpha_u)\Sigma_1^{-1} + \alpha_u\Sigma_2^{-1})^{-1}$ and $\mu_{\alpha_u}=\Sigma_{\alpha_u}((1-\alpha_u)\Sigma_1^{-1}\mu_1 + \alpha_u\Sigma_2^{-1}\mu_2)$,  $D$ denotes the number of dimension, and $det[\cdot]$ represents the determinant.}
\smallskip

With Theorem 1, one can calculate $JS^{G_{\alpha_u}}$ or its dual form $JS_*^{G_{\alpha_u}}$ based on the mean vector and the variance vector, and use $JS^{G_{\alpha_u}}$ or $JS_*^{G_{\alpha_u}}$ to replace the original KL divergence term in the ELBO (Eq.~(\ref{eq:elbo})) for training the whole framework. When the two distributions are diagonal Gaussians, $\Sigma_1$ and $\Sigma_2$ can be implemented by diagonal matrices with the variance vectors for calculating $JS^{G_{\alpha_u}}$ or~$JS_*^{G_{\alpha_u}}$. 

\subsubsection{Loss Functions} 



To calculate loss functions, reliable pseudo-labels are required for unlabeled data. In practice, to select reliable unlabeled samples from $\mathcal{U}$ and  their corresponding pseudo-labels, we preset a confidence threshold ($\tau_c$) and an uncertainty threshold ($\tau_u$). In particular, as for unlabeled data $x_{i}^{u}$, NP-Match gives its prediction $p(y|Aug_w(x_{i}^{u}))$ and associated uncertainty estimate under the inference mode, where $Aug_w(\cdot)$ denotes the weak augmentation. When the highest prediction score $max(p(y|Aug_w(x_{i}^{u})))$ is higher than $\tau_c$, and the uncertainty is smaller than $\tau_u$, the sample will be chosen, and we denote the selected sample as $x_{i}^{u_c}$, since the model is certain about his prediction, and the pseudo-label of $x_{i}^{u_c}$ is $\hat y_i= arg\ max(p(y|Aug_w(x_{i}^{u_c})))$. 
Concerning $\mu B$ unlabeled samples in $\mathcal{U}$, we assume $B_c$ unlabeled samples are selected from them in each feedforward pass. 
According to the ELBO (Eq.~(\ref{eq:elbo})), three loss terms are used for training, namely, $L_{cls}$, $L^u_{cls}$, and $JS^{G_{\alpha_u}}$. 
For each input (labeled or unlabeled), the NP model can give $T$ predictions, and hence~$L_{cls}$ and $L^u_{cls}$ are defined as:
\begin{equation}
\small
\begin{aligned}
\label{eq:cross-entropy}
&L_{cls} = \frac{1}{B \times T} \sum^{B}_{i=1} \sum^{T}_{j=1} H(y^*_i, p_j(y|Aug_w(x_i))), \\
&L^u_{cls} = \frac{1}{B_c \times T} \sum^{B_c}_{i=1} \sum^{T}_{j=1} H(\hat y_i, p_j(y|Aug_s(x_{i}^{u_{c}}))), 
\end{aligned}
\end{equation} 
where $Aug_s(\cdot)$ denotes the strong augmentation, $y^*_i$ represents the real label for the labeled sample $x_{i}$, and $H(\cdot, \cdot)$ denotes the cross-entropy between two distributions. Thus, the total loss function is given by:
\begin{equation}
\label{eq:overall}
L_{total} = L_{\text{cls}} + \lambda_u L^u_{cls} + \beta JS^{G_{\alpha_u}},
\end{equation}
where $\lambda_u$ and $\beta$ are coefficients. During training, we followed previous work \cite{sohn2020fixmatch,zhang2021flexmatch, rizve2021defense, li2021comatch} to utilize  the exponential moving average (EMA) technique. It is worth noting that, in the real implementation, NP-Match only preserves the averaged representation over all representations in each memory bank after training, which just takes up negligible storage space.

\begin{table*}
\centering 
\resizebox{\textwidth}{!}{
\begin{tabular}{@{}cccccccccc@{}}
 \toprule[1pt]
Dataset  & \multicolumn{3}{c}{CIFAR-10} &\multicolumn{3}{c}{CIFAR-100}  & \multicolumn{3}{c}{STL-10}\\
 \hline  
Label Amount & 40 & 250 & 4000 & 400 & 2500 & 10000 & 40 & 250 & 1000\\
 \hline 
  \quad MixMatch \cite{berthelot2019mixmatch} &  36.19 {\small($\pm$6.48)} &  13.63 {\small($\pm$0.59)} &  6.66 {\small($\pm$0.26)} &  67.59 {\small($\pm$0.66)} & 39.76 {\small($\pm$0.48)}  & 27.78 {\small($\pm$0.29)} &  54.93 {\small($\pm$0.96)} &  34.52 {\small($\pm$0.32)} &  21.70 {\small($\pm$0.68)} \\
 \quad ReMixMatch \cite{berthelot2019remixmatch}  &  9.88 {\small($\pm$1.03)} &  6.30 {\small($\pm$0.05)} &  4.84 {\small($\pm$0.01)} &  42.75 {\small($\pm$1.05)} & {\bf 26.03} {\small($\pm$0.35)}  & {\bf 20.02} {\small($\pm$0.27)} &  32.12 {\small($\pm$6.24)} &  12.49 {\small($\pm$1.28)} &  6.74 {\small($\pm$0.14)} \\
 \quad UDA \cite{xie2019unsupervised}  &  10.62 {\small($\pm$3.75)} &  5.16 {\small($\pm$0.06)} &  4.29 {\small($\pm$0.07)} &  46.39 {\small($\pm$1.59)}  & 27.73 {\small($\pm$0.21)} & 22.49 {\small($\pm$0.23)}  & 37.42 {\small($\pm$8.44)} &  9.72 {\small($\pm$1.15)} &  6.64 {\small($\pm$0.17)} \\
 \quad CoMatch \cite{li2021comatch}  &   6.88 {\small($\pm$0.92)} &  4.90 {\small($\pm$0.35)} &  4.06 {\small($\pm$0.03)} &  40.02 {\small($\pm$1.11)}  & 27.01 {\small($\pm$0.21)} & 21.83 {\small($\pm$0.23)}  & 31.77 {\small($\pm$2.56)} & 11.56 {\small($\pm$1.27)} &   8.66  {\small($\pm$0.41)} \\
 \quad SemCo \cite{nassar2021all}  &   7.87 {\small($\pm$0.22)} &   5.12 {\small($\pm$0.27)} &  {\bf 3.80} {\small($\pm$0.08)} &  44.11 {\small($\pm$1.18)}  & 31.93 {\small($\pm$0.33)} & 24.45 {\small($\pm$0.12)}  & 34.17 {\small($\pm$2.78)} & 12.23 {\small($\pm$1.40)} &   7.49  {\small($\pm$0.29)} \\
  \quad Meta Pseudo Labels \cite{pham2021meta}  &   6.93 {\small($\pm$0.17)} &   4.94 {\small($\pm$0.04)} &  3.89 {\small($\pm$0.07)} &  44.23 {\small($\pm$0.99)}  & 27.68 {\small($\pm$0.22)} & 22.48 {\small($\pm$0.18)}  & 34.29 {\small($\pm$3.29)} & 9.90 {\small($\pm$0.96)} &   6.45  {\small($\pm$0.26)} \\
 \quad FlexMatch \cite{zhang2021flexmatch} &   4.96 {\small($\pm$0.06)} &  4.98 {\small($\pm$0.09)} &  4.19 {\small($\pm$0.01)} &  39.94 {\small($\pm$1.62)}  & 26.49 {\small($\pm$0.20)} & 21.90 {\small($\pm$0.15)}  & 29.15 {\small($\pm$4.16)} & {\bf 8.23} {\small($\pm$0.39)} &   5.77  {\small($\pm$0.18)} \\
 \quad UPS \cite{rizve2021defense}  &  5.26 {\small($\pm$0.29)} &  5.11 {\small($\pm$0.08)} &  4.25 {\small($\pm$0.05)} &  41.07 {\small($\pm$1.66)}  & 27.14 {\small($\pm$0.24)} & 21.97 {\small($\pm$0.23)}  & 30.82  {\small($\pm$2.16)} & 9.77 {\small($\pm$0.44)} &   6.02  {\small($\pm$0.28)} \\
 \quad FixMatch \cite{sohn2020fixmatch}  &   7.47 {\small($\pm$0.28)} &  {\bf 4.86} {\small($\pm$0.05)} &  4.21 {\small($\pm$0.08)} &  46.42 {\small($\pm$0.82)}  & 28.03 {\small($\pm$0.16)} & 22.20 {\small($\pm$0.12)}  & 35.96 {\small($\pm$4.14)} & 9.81 {\small($\pm$1.04)} &   6.25  {\small($\pm$0.33)} \\
 \quad NP-Match (ours) &  {\bf 4.91} {\small($\pm$0.04)} &  4.96 {\small($\pm$0.06)} &  4.11 {\small($\pm$0.02)} & {\bf 38.91} {\small($\pm$0.99)}  & {\bf 26.03} {\small($\pm$0.26)} & 21.22 {\small($\pm$0.13)}  & {\bf 14.20} {\small($\pm$0.67)} & 9.51 {\small($\pm$0.37)} & {\bf 5.59}  {\small($\pm$0.24)}\\
  \bottomrule[1pt]
  \end{tabular}} 
 \caption{Comparison with SOTA results on CIFAR-10, CIFAR-100, and STL-10. The error rates are reported with standard deviation. } 
 \label{tab:compare_sota} 
 \end{table*}
 
 \begin{figure*}[t]
\centering
\includegraphics[width=\linewidth]{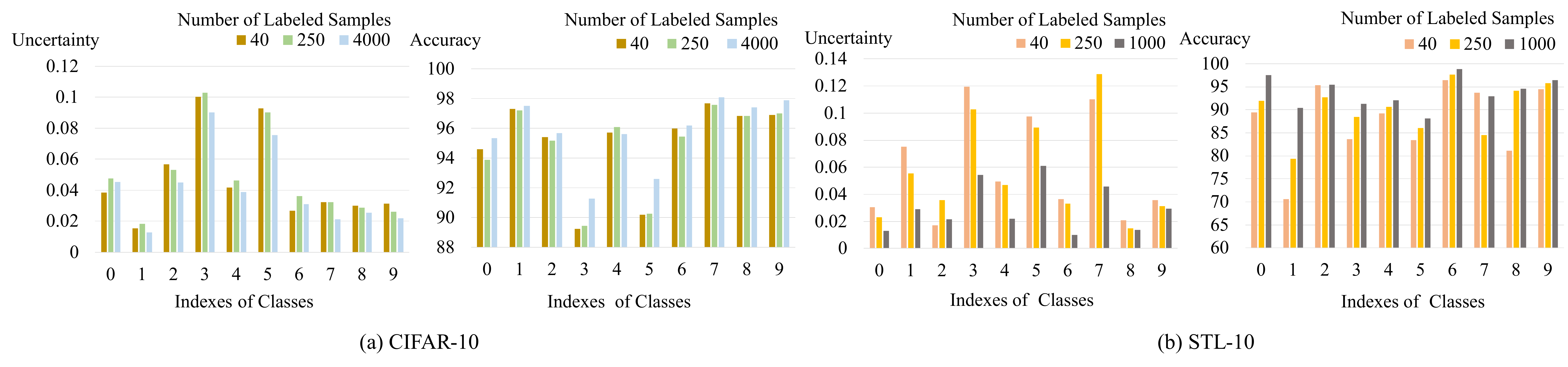}
\vspace{-5ex}
\caption{Analysis of average class-wise uncertainty and accuracy.}
\label{fig:u_vs_p} 
\end{figure*}
 
\section{Experiments}
\label{sec:experiment}
We now report on our experiments of NP-Match on four public image classification benchmarks. To save space, implementation details are given in the appendix. 

\subsection{Datasets}
We conducted our experiments on four widely used public SSL benchmarks, including CIFAR-10 \cite{krizhevsky2009learning}, CIFAR-100 \cite{krizhevsky2009learning}, STL-10 \cite{coates2011analysis}, and ImageNet \cite{deng2009imagenet}. CIFAR-10 and CIFAR-100 contain 50,000 images of size $32\times32$ from 10 and 100 classes, respectively. We evaluated NP-match on these two datasets following the evaluation settings used in previous works \cite{sohn2020fixmatch, zhang2021flexmatch, li2021comatch}.
The STL-10 dataset has 5000 labeled samples with size $96\times96$ from 10 classes and 100,000 unlabeled samples, and it is more difficult than CIFAR,  since STL-10  has a number of out-of-distribution images in the unlabeled set. We follow the experimental settings for STL-10 as detailed in \cite{zhang2021flexmatch}. Finally, ImageNet contains around 1.2 million images from 1000 classes. Following the experimental settings in \cite{zhang2021flexmatch}, we used 100K labeled data, namely, 100 labels per class. 

\begin{table}
\centering 
\resizebox{0.45\textwidth}{!}{
\begin{tabular}{@{}ccccccc @{}}
 \toprule[1pt]
Dataset  & \multicolumn{3}{c}{CIFAR-10} &\multicolumn{3}{c}{STL-10}   \\
 \hline  
Label Amount & 40 & 250 & 4000 & 40 & 250 & 1000 \\
 \hline 
 UPS (MC Dropout) & 7.96  & 7.02   & {\bf 5.82}  & 17.23  &  9.65  & 5.69 \\
 NP-Match & {\bf 7.23}  & {\bf 6.85}  &  5.89 &  {\bf 12.45}  &  {\bf 8.72}   &  {\bf 5.28} \\
  \bottomrule[1pt]
  \end{tabular}}\vspace*{-1ex}
 \caption{Expected UCEs (\%) of the MC-dropout-based model (i.e., UPS \cite{rizve2021defense}) and of NP-Match on the test sets of CIFAR-10 and STL-10.}  
 \label{tab:ablation_uce} 
 \end{table}

\subsection{Main Results}
\label{sec:main_results} 
In the following, we report the main experimental results on the accuracy, 
the average uncertainty, the expected uncertainty calibration error, and the running time of NP-Match compared with SOTA approaches.

First, in Table~\ref{tab:compare_sota}, we compare NP-Match with SOTA SSL image classification methods on CIFAR-10, CIFAR-100, and STL-10. We see that NP-Match outperforms SOTA results or achieves competitive results under different SSL settings. We highlight two key observations. First, NP-Match outperforms all other methods by a wide margin on all three benchmarks under the most challenging settings,  where the number of labeled samples is smallest.
Second, NP-Match is compared to  UPS\footnote{Note that UPS \cite{rizve2021defense} does not use strong augmentations, thus we re-implemented it with RandAugment \cite{cubuk2020randaugment} for fair comparisons.}, since the UPS framework is the MC-dropout-based probabilistic model for semi-supervised image classification, and NP-Match completely outperforms them on all three benchmarks. This suggests that NPs can be a good alternative to MC dropout in probabilistic approaches to semi-supervised learning tasks.
 
 \begin{figure}[t]
\centering
\includegraphics[width=\linewidth]{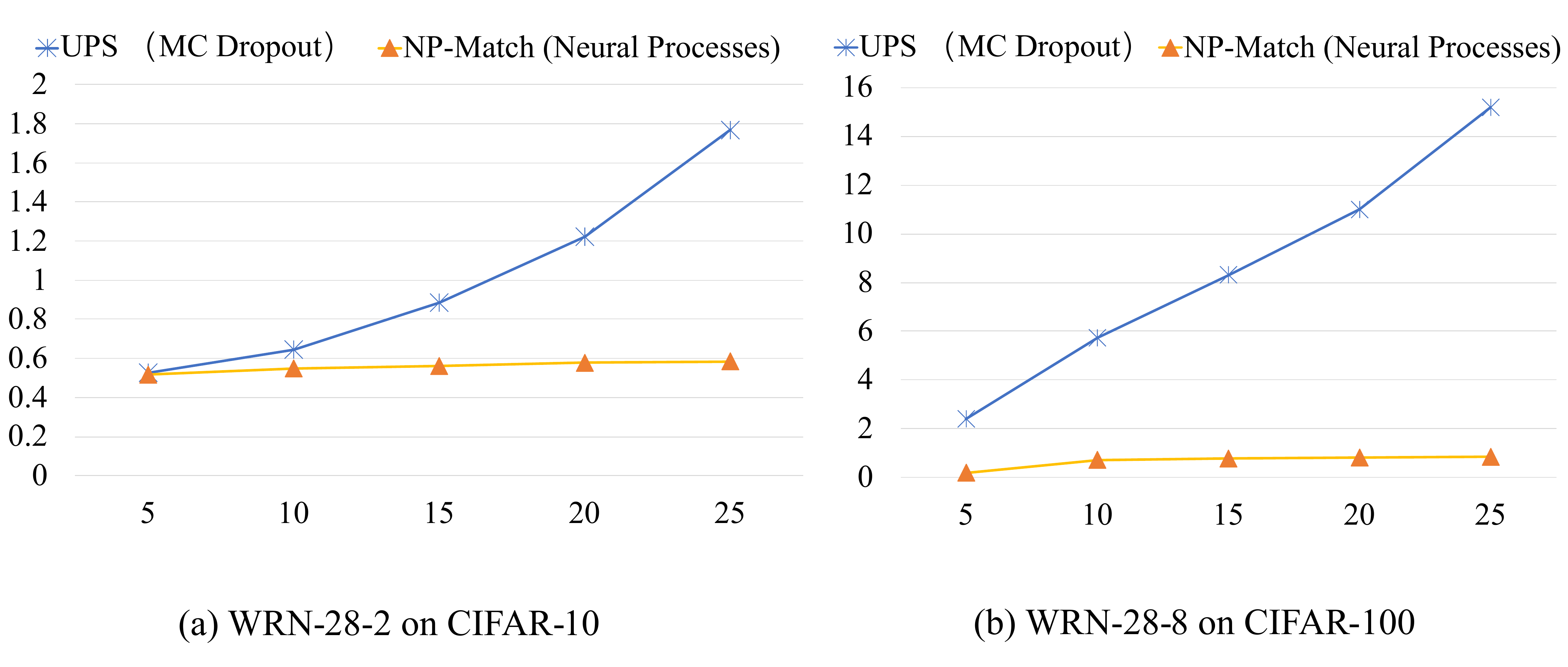}
\vspace{-5ex}
\caption{Time consumption of estimating uncertainty for the MC-dropout-based model (i.e., UPS \cite{rizve2021defense}) and  NP-Match. The horizontal axis refers to the number of predictions used for the uncertainty quantification, and the vertical axis indicates the time consumption (sec).} 
\label{fig:speed}
\end{figure}

\begin{table}
\centering 
\resizebox{0.4\textwidth}{!}{
\begin{tabular}{@{}c|ccc@{}}
 \toprule[1pt]
   & Method & Top-1  & Top-5 \\
 \hline 
  \multirow{3}{*}{\shortstack{Deterministic \\ Methods}} & FixMatch \cite{sohn2020fixmatch} &   43.66  &  21.80 \\ 
 & FlexMatch \cite{zhang2021flexmatch} &  41.85  & 19.48 \\ 
  &CoMatch \cite{li2021comatch}&  42.17 &  19.64 \\
 \hline
\multirow{2}{*}{\shortstack{Probabilistic \\ Methods}} &UPS \cite{rizve2021defense} &  42.69  &  20.23 \\
 & NP-Match  &  {\bf 41.78} &  {\bf 19.33} \\ 
  \bottomrule[1pt]
  \end{tabular}} 
 \caption{Error rates of SOTA methods on ImageNet.} \vspace*{-2ex}
 \label{tab:compare_imagenet} 
 \end{table}

\begin{table*}
\centering 
\resizebox{\textwidth}{!}{
\begin{tabular}{@{}cccccccccc@{}}
 \toprule[1pt]
Dataset  & \multicolumn{3}{c}{CIFAR-10} &\multicolumn{3}{c}{CIFAR-100}  & \multicolumn{3}{c}{STL-10}\\
 \hline  
Label Amount & 40 & 250 & 4000 & 400 & 2500 & 10000 & 40 & 250 & 1000\\
 \hline 
 NP-Match with KL &  5.32  {\small($\pm$0.06)} &  5.20 {\small($\pm$0.02)} & 4.36 {\small($\pm$0.03)} & 39.15 {\small($\pm$1.53)} &  26.48  {\small($\pm$0.23)} & 21.51 {\small($\pm$0.17)} &   14.67 {\small($\pm$0.38)}   & 9.92 {\small($\pm$0.24)}  & 6.21  {\small($\pm$0.23)} \\
 \quad NP-Match with $JS_*^{G_{\alpha_u}}$  &  4.93  {\small($\pm$0.02)} &  {\bf 4.87} {\small($\pm$0.03)} & 4.19 {\small($\pm$0.04)} & {\bf 38.67} {\small($\pm$1.29)} &  26.24  {\small($\pm$0.17)} & 21.33 {\small($\pm$0.10)} &   14.45 {\small($\pm$0.55)}   & {\bf 9.48} {\small($\pm$0.28)}  & {\bf 5.47}  {\small($\pm$0.19)} \\
 \quad NP-Match with $JS^{G_{\alpha_u}}$  & {\bf 4.91} {\small($\pm$0.04)} &  4.96 {\small($\pm$0.06)} &  {\bf 4.11} {\small($\pm$0.02)} &  38.91 {\small($\pm$0.99)}  & {\bf 26.03} {\small($\pm$0.26)} & {\bf 21.22} {\small($\pm$0.13)}  & {\bf 14.20} {\small($\pm$0.67)} & 9.51 {\small($\pm$0.37)} &  5.59  {\small($\pm$0.24)}\\ 
  \bottomrule[1pt]
  \end{tabular}}\vspace*{-1ex}
 \caption{Ablation studies of the proposed uncertainty-guided skew-geometric JS divergence and its dual form.} 
 \label{tab:ablation_jsd} 
 \end{table*}

Second, we analyse the relationship between the average class-wise uncertainty and accuracy at test phase on CIFAR-10 and STL10. From Figure~\ref{fig:u_vs_p}, we empirically observe that: (1) when more labeled data are used for training, the average uncertainty of samples' predictions for each class decreases. This is consistent with the property of NPs and GPs where the model is less uncertain with regard to its prediction when more real and correct labels are leveraged; (2) the classes with higher average uncertainties have lower accuracy, meaning that the uncertainty is a good standard for choosing unlabeled samples.

Third, the expected uncertainty calibration error (UCE) of our method is also calculated to evaluate the uncertainty estimation. The expected UCE is used to measure the miscalibration of uncertainty \cite{laves2020calibration}, which is an analogue to the expected calibration error (ECE) \cite{guo2017calibration, naeini2015obtaining}. The low expected UCE indicates that the model is certain when making accurate predictions and that the model is uncertain when making inaccurate predictions. More details about the expected UCE can be found in previous works \cite{laves2020calibration, krishnan2020improving}. The results of NP-Match and the MC-dropout-based model (i.e., UPS \cite{rizve2021defense}) are shown in Table~\ref{tab:ablation_uce}; their comparison shows that NP-Match can output more reliable and well-calibrated uncertainty estimates. 

Furthermore, we compare the running time of NP-Match and the MC dropout-based model (i.e., UPS  \cite{rizve2021defense}). We use a batch of 16 samples and two network architectures that are widely used in previous works \cite{zagoruyko2016wide, zhang2021flexmatch, sohn2020fixmatch, li2021comatch}, namely, WRN-28-2 on CIFAR-10 (Figure~\ref{fig:speed} (a)) and WRN-28-8 on CIFAR-100 (Figure~\ref{fig:speed} (b)).
In (a), we observe that when the number of predictions ($T$) increases,  the time cost of the UPS framework rises quickly, but the time cost of NP-Match grows slowly. In (b), we observe that the time cost gap between these two methods is even larger when a larger model is tested on a larger dataset. This demonstrates that NP-Match is significantly more computationally efficient than MC dropout-based methods. 
 
Finally, Table~\ref{tab:compare_imagenet} shows the experiments conducted on ImageNet. Here, NP-Match achieves a SOTA  performance, suggesting that it is effective at handling challenging large-scale datasets. Note that previous works usually evaluate their frameworks under distinct SSL settings, and thus it is hard to compare different methods directly. Therefore, we re-evaluate another two methods proposed recently under the same SSL setting with the same training details, namely, UPS and CoMatch. 
 
 \subsection{Ablation Studies}
We now report our ablation studies on CIFAR-10, CIFAR-100, and STL-10. We also present further experiments, including a hyperparameter exploration, in the appendix.

We evaluate our uncertainty-guided skew-geometric JS divergence ($JS^{G_{\alpha_u}}$) as well as its dual form ($JS_*^{G_{\alpha_u}}$), and compare them to the original KL divergence in NPs. In Table~\ref{tab:ablation_jsd}, we see that NP-Match with KL divergence consistently underperforms relative to our proposed $JS^{G_{\alpha_u}}$ and $JS_*^{G_{\alpha_u}}$. This suggests that our uncertainty-guided skew-geometric JS divergence can mitigate the problem caused by low-quality feature representations. Between the two, $JS^{G_{\alpha_u}}$ and $JS_*^{G_{\alpha_u}}$ achieve a comparable performance across the three benchmarks, and thus  we select $JS^{G_{\alpha_u}}$ to replace the original KL divergence in the ELBO (Eq.~(\ref{eq:elbo})) for the comparisons to previous SOTA methods in Section~\ref{sec:main_results}.

\section{Summary and Outlook}
\label{sec:conclusion}
In this work, we proposed the application of neural processes (NPs) to semi-supervised learning (SSL), designing a new framework called NP-Match, and explored its use in semi-supervised large-scale image classification. 
To our knowledge, this is the first such work. To better adapt NP-Match to the SSL task, we proposed a new divergence term, which we call uncertainty-guided skew-geometric JS divergence, to replace the original  KL divergence in NPs. We demonstrated the effectiveness of NP-Match and the proposed divergence term for SSL in extensive experiments, and also showed that NP-Match could be a good alternative~to MC dropout in SSL. 

Future works will explore the following two directions.
First, due to the successful application of NPs to semi-supervised image classification, it is valuable to explore NPs in other SSL tasks, such as object detection and segmentation.
Second, many successful NPs variants have been proposed since the original NPs \cite{garnelo2018neural} (see Section~\ref{sec:related_work}). We will  also explore these in SSL for image classification. 

\section{Acknowledgements}
This work was partially supported by the Alan Turing Institute under the EPSRC grant EP/N510129/1, by the AXA Research Fund, and by the EPSRC grant EP/R013667/1. We also acknowledge the use of the EPSRC-funded Tier 2 facility JADE (EP/P020275/1) and GPU computing support by Scan Computers International Ltd.


\bibliography{example_paper}
\bibliographystyle{icml2022}

\newpage
\appendix
\onecolumn

\icmltitle{Appendix}

\section{Derivation of ELBO (Eq.~(\ref{eq:elbo}))}

\emph{Proof.} As for the marginal joint distribution $p(y_{1:n} | x_{1:n})$ over $n$ data points in which there are $m$ context points and $r$ target points (i.e., $m+r=n$), we assume a variation distribution $q$, and then: 
\begin{equation}
\footnotesize		
\begin{aligned} 
 &log \ p(y_{1:n} | x_{1:n}) = log \int_z p(z, y_{1:n} | x_{1:n}) \\
 & = log \int_z \frac{p(z, y_{1:n}|x_{1:n})}{q(z|x_{m+1:\ m+r}, y_{m+1:\ m+r})}q(z|x_{m+1:\ m+r}, y_{m+1:\ m+r}) \\
 & \ge \mathbb{E}_{q(z|x_{m+1:\ m+r}, y_{m+1:\ m+r})}[log \ \frac{p(z, y_{1:n}|x_{1:n})}{q(z|x_{m+1:\ m+r}, y_{m+1:\ m+r})}] \\
 & = \mathbb{E}_{q(z|x_{m+1:\ m+r}, y_{m+1:\ m+r})}[log \ \frac{p(y_{1:m})p(z|x_{1:m}, y_{1:m})\prod^{m+r}_{i=m+1}p(y_i|z, x_i)}{q(z|x_{m+1:\ m+r}, y_{m+1:\ m+r})}] \\
 & = \mathbb{E}_{q(z|x_{m+1:\ m+r}, y_{m+1:\ m+r})}[\sum^{m+r}_{i=m+1} log \ p(y_i|z, x_i) + log \ \frac{p(z|x_{1:m}, y_{1:m})}{q(z|x_{m+1:\ m+r}, y_{m+1:\ m+r})} + log \ p(y_{1:m})] \\
 & = \mathbb{E}_{q(z|x_{m+1:\ m+r}, y_{m+1:\ m+r})}[\sum^{m+r}_{i=m+1} log \ p(y_i|z, x_i) - log \ \frac{q(z|x_{m+1:\ m+r}, y_{m+1:\ m+r})}{p(z|x_{1:m}, y_{1:m})}] + const,
\end{aligned}
\end{equation}
where ``$const$'' refers to $\mathbb{E}_{q(z|x_{m+1:\ m+r}, y_{m+1:\ m+r})}[log \ p(y_{1:m})]$, which is a constant term.  Concerning that $p(z|x_{1:m}, y_{1:m})$ is unknown, we replace it with $q(z|x_{1:m}, y_{1:m})$, and then we get:
\begin{equation}
\small
\begin{aligned} 
&log\ p(y_{1:n}|x_{1:n}) \ge \mathbb{E}_{q(z|x_{m+1:\ m+r}, y_{m+1:\ m+r})}\Big[\sum^{m+r}_{i=m+1}log\ p(y_i|z, x_i) - log\ \frac{q(z|x_{m+1:\ m+r}, y_{m+1:\ m+r})}{q(z|x_{1:m}, y_{1:m})}\Big] + const.
\end{aligned}
\end{equation}
\hfill $\square$

\section{Proof of Theorem 1}

\emph{Proof.} Let us first show that  $\Sigma_{\alpha_u}=((1-\alpha_u)\Sigma_1^{-1} + \alpha_u\Sigma_2^{-1})^{-1}$ and $\mu_{\alpha_u}=\Sigma_{\alpha_u}((1-\alpha_u)\Sigma_1^{-1}\mu_1 + \alpha_u\Sigma_2^{-1}\mu_2)$. Concerning two Gaussian distributions $\mathcal{N}_1(\mu_1, \Sigma_1)$ and $\mathcal{N}_2(\mu_2, \Sigma_2)$, the weighted geometric mean of them ($\mathcal{N}_1^{1-\alpha_u}\mathcal{N}_2^{\alpha_u}$) is given by:
\begin{equation}
\footnotesize 
\begin{aligned}
&(2\pi)^{-\frac{D}{2}}det[\Sigma_1]^{-\frac{1-\alpha_u}{2}}det[\Sigma_2]^{-\frac{\alpha_u}{2}}e^{-\frac{1-\alpha_u}{2}(x-\mu_1)^T\Sigma_1^{-1}(x-\mu_1) - \frac{\alpha_u}{2}(x-\mu_2)^T\Sigma_2^{-1}(x-\mu_2)} \\
&=(2\pi)^{-\frac{D}{2}}det[\Sigma_1]^{-\frac{1-\alpha_u}{2}}det[\Sigma_2]^{-\frac{\alpha_u}{2}}e^{-\frac{1}{2}((x-\mu_1)^T((1-\alpha_u)\Sigma_1^{-1})(x-\mu_1) + (x-\mu_2)^T(\alpha_u\Sigma_2^{-1})(x-\mu_2))}.
\end{aligned}
\end{equation}
Now, we let $\Sigma_{1_u}^{-1} = (1-\alpha_u)\Sigma_1^{-1}$ and $\Sigma_{2_u}^{-1} = \alpha_u\Sigma_2^{-1}$, then:
\begin{equation}
\footnotesize		
\label{eqn:proof1}
\begin{aligned}
&(2\pi)^{-\frac{D}{2}}det[\Sigma_1]^{-\frac{1-\alpha_u}{2}}det[\Sigma_2]^{-\frac{\alpha_u}{2}}e^{-\frac{1}{2}((x-\mu_1)^T\Sigma_{1_u}^{-1}(x-\mu_1)+(x-\mu_2)^T\Sigma_{2_u}^{-1}(x-\mu_2))} \\
&= C_1e^{-\frac{1}{2}(x^T(\Sigma_{1_u}^{-1} + \Sigma_{2_u}^{-1})x - x^T(\Sigma_{1_u}^{-1}\mu_1 + \Sigma_{2_u}^{-1}\mu_2)-(\mu_1^T\Sigma_{1_u}^{-1}+\mu_2^T\Sigma_{2_u}^{-1})x + (\mu_1^T \Sigma_{1_u}^{-1} \mu_1 + \mu_2^T \Sigma_{2_u}^{-1}\mu_2))} \\
&=  C_1e^{-\frac{1}{2}(x^T(\Sigma_{1_u}^{-1} + \Sigma_{2_u}^{-1})x - x^T(\Sigma_{1_u}^{-1} + \Sigma_{2_u}^{-1})(\Sigma_{1_u}^{-1} + \Sigma_{2_u}^{-1})^{-1}(\Sigma_{1_u}^{-1}\mu_1 + \Sigma_{2_u}^{-1}\mu_2)-(\mu_1^T\Sigma_{1_u}^{-1}+\mu_2^T\Sigma_{2_u}^{-1})x + (\mu_1^T \Sigma_{1_u}^{-1} \mu_1 + \mu_2^T \Sigma_{2_u}^{-1}\mu_2))} \\
&=  C_1e^{-\frac{1}{2}(x^T(\Sigma_{1_u}^{-1} + \Sigma_{2_u}^{-1})(x - (\Sigma_{1_u}^{-1} + \Sigma_{2_u}^{-1})^{-1}(\Sigma_{1_u}^{-1}\mu_1 + \Sigma_{2_u}^{-1}\mu_2))-(\mu_1^T\Sigma_{1_u}^{-1}+\mu_2^T\Sigma_{2_u}^{-1})x + (\mu_1^T \Sigma_{1_u}^{-1} \mu_1 + \mu_2^T \Sigma_{2_u}^{-1}\mu_2))} 
\end{aligned}
\end{equation}

\begin{equation}
\footnotesize	
\begin{aligned}
&=  C_1e^{-\frac{1}{2}(x^T(\Sigma_{1_u}^{-1} + \Sigma_{2_u}^{-1})(x - (\Sigma_{1_u}^{-1} + \Sigma_{2_u}^{-1})^{-1}(\Sigma_{1_u}^{-1}\mu_1 + \Sigma_{2_u}^{-1}\mu_2))-(\mu_1^T\Sigma_{1_u}^{-1}+\mu_2^T\Sigma_{2_u}^{-1})x + (\mu_1^T\Sigma_{1_u}^{-1}+\mu_2^T\Sigma_{2_u}^{-1})(\Sigma_{1_u}^{-1} + \Sigma_{2_u}^{-1})^{-1}(\Sigma_{1_u}^{-1}\mu_1 + \Sigma_{2_u}^{-1}\mu_2) + C_2)} \\
&=  C_1e^{-\frac{1}{2}(x^T(\Sigma_{1_u}^{-1} + \Sigma_{2_u}^{-1})(x - (\Sigma_{1_u}^{-1} + \Sigma_{2_u}^{-1})^{-1}(\Sigma_{1_u}^{-1}\mu_1 + \Sigma_{2_u}^{-1}\mu_2))-(\mu_1^T\Sigma_{1_u}^{-1}+\mu_2^T\Sigma_{2_u}^{-1})(x - (\Sigma_{1_u}^{-1} + \Sigma_{2_u}^{-1})^{-1}(\Sigma_{1_u}^{-1}\mu_1 + \Sigma_{2_u}^{-1}\mu_2)) + C_2)} \\
&=  C_1e^{-\frac{1}{2}(x^T(\Sigma_{1_u}^{-1} + \Sigma_{2_u}^{-1})(x - (\Sigma_{1_u}^{-1} + \Sigma_{2_u}^{-1})^{-1}(\Sigma_{1_u}^{-1}\mu_1 + \Sigma_{2_u}^{-1}\mu_2))-(\mu_1^T\Sigma_{1_u}^{-1}+\mu_2^T\Sigma_{2_u}^{-1})(\Sigma_{1_u}^{-1} + \Sigma_{2_u}^{-1})^{-1}(\Sigma_{1_u}^{-1} + \Sigma_{2_u}^{-1})(x - (\Sigma_{1_u}^{-1} + \Sigma_{2_u}^{-1})^{-1}(\Sigma_{1_u}^{-1}\mu_1 + \Sigma_{2_u}^{-1}\mu_2)) + C_2)} \\
&=  C_1e^{-\frac{1}{2}((x^T - (\mu_1^T\Sigma_{1_u}^{-1}+\mu_2^T\Sigma_{2_u}^{-1})(\Sigma_{1_u}^{-1} + \Sigma_{2_u}^{-1})^{-1})(\Sigma_{1_u}^{-1} + \Sigma_{2_u}^{-1})(x - (\Sigma_{1_u}^{-1} + \Sigma_{2_u}^{-1})^{-1}(\Sigma_{1_u}^{-1}\mu_1 + \Sigma_{2_u}^{-1}\mu_2))+ C_2)} \\
&= C_1e^{-\frac{1}{2}((x - (\Sigma_{1_u}^{-1} + \Sigma_{2_u}^{-1})^{-1}(\Sigma_{1_u}^{-1}\mu_1 + \Sigma_{2_u}^{-1}\mu_2))^T(\Sigma_{1_u}^{-1} + \Sigma_{2_u}^{-1})(x - (\Sigma_{1_u}^{-1} + \Sigma_{2_u}^{-1})^{-1}(\Sigma_{1_u}^{-1}\mu_1 + \Sigma_{2_u}^{-1}\mu_2))+ C_2)} \\
&= C_3e^{-\frac{1}{2}(x - (\Sigma_{1_u}^{-1} + \Sigma_{2_u}^{-1})^{-1}(\Sigma_{1_u}^{-1}\mu_1 + \Sigma_{2_u}^{-1}\mu_2))^T(\Sigma_{1_u}^{-1} + \Sigma_{2_u}^{-1})(x - (\Sigma_{1_u}^{-1} + \Sigma_{2_u}^{-1})^{-1}(\Sigma_{1_u}^{-1}\mu_1 + \Sigma_{2_u}^{-1}\mu_2))}, 
\end{aligned}
\end{equation}
where $C_1 = (2\pi)^{-\frac{D}{2}}det[\Sigma_1]^{-\frac{1-\alpha_u}{2}}det[\Sigma_2]^{-\frac{\alpha_u}{2}}$, $C_2$ is a constant for aborting the terms used for completing the square relative to $x$, and $C_3 = C_1e^{-\frac{1}{2}C_2}$.
The last formula of Eq.~(\ref{eqn:proof1}) is an unnormalized Gaussian curve with covariance $(\Sigma_{1_u}^{-1} + \Sigma_{2_u}^{-1})^{-1}$ and mean $(\Sigma_{1_u}^{-1} + \Sigma_{2_u}^{-1})^{-1}(\Sigma_{1_u}^{-1}\mu_1 + \Sigma_{2_u}^{-1}\mu_2)$. Therefore, we can get $\Sigma_{\alpha_u}=((1-\alpha_u)\Sigma_1^{-1} + \alpha_u\Sigma_2^{-1})^{-1}$ and $\mu_{\alpha_u}=\Sigma_{\alpha_u}((1-\alpha_u)\Sigma_1^{-1}\mu_1 + \alpha_u\Sigma_2^{-1}\mu_2)$. After the normalization step, we can get a Gaussian distribution $\mathcal{N}_{\alpha_u}(\mu_{\alpha_u}, \Sigma_{\alpha_u})$.

\smallskip
As for $JS^{G_{\alpha_u}}$, we first calculate $\mathbb{E}_{\mathcal{N}_1}[log{\mathcal{N}_1} - log{\mathcal{N}_{\alpha_u}}]$ as follows:
\begin{equation}
\footnotesize		 
\begin{aligned}
&\mathbb{E}_{\mathcal{N}_1}[log{\mathcal{N}_1} - log{\mathcal{N}_{\alpha_u}}] \\
&= \frac{1}{2} \mathbb{E}_{\mathcal{N}_1}[-logdet[\Sigma_1] - (x-\mu_1)^T\Sigma_1^{-1}(x-\mu_1) + logdet[\Sigma_{\alpha_u}] + (x-\mu_{\alpha_u})^T\Sigma_{\alpha_u}^{-1}(x-\mu_{\alpha_u}) ]  \\
&= \frac{1}{2} ( log\frac{det[\Sigma_{\alpha_u}]}{det[\Sigma_1]} + \mathbb{E}_{\mathcal{N}_1}[ - (x-\mu_1)^T\Sigma_1^{-1}(x-\mu_1) +  (x-\mu_{\alpha_u})^T\Sigma_{\alpha_u}^{-1}(x-\mu_{\alpha_u}) ] ) \\
&= \frac{1}{2} ( log\frac{det[\Sigma_{\alpha_u}]}{det[\Sigma_1]} + \mathbb{E}_{\mathcal{N}_1}[ - tr[\Sigma_1^{-1}\Sigma_1] +  tr[\Sigma_{\alpha_u}^{-1}(xx^T-2x\mu_{\alpha_u}^T+\mu_{\alpha_u}\mu_{\alpha_u}^T)] ] ) \\
&= \frac{1}{2} log\frac{det[\Sigma_{\alpha_u}]}{det[\Sigma_1]} - \frac{D}{2}  + \frac{1}{2} \mathbb{E}_{\mathcal{N}_1}[  tr[\Sigma_{\alpha_u}^{-1}(xx^T-2x\mu_{\alpha_u}^T+\mu_{\alpha_u}\mu_{\alpha_u}^T)] ]  \\
&= \frac{1}{2} log\frac{det[\Sigma_{\alpha_u}]}{det[\Sigma_1]} - \frac{D}{2}  + \frac{1}{2} \mathbb{E}_{\mathcal{N}_1}[  tr[\Sigma_{\alpha_u}^{-1}((x-\mu_1)(x-\mu_1)^T + 2\mu_1x^T-\mu_1\mu_1^T-2x\mu_{\alpha_u}^T+\mu_{\alpha_u}\mu_{\alpha_u}^T)] ]  \\
&= \frac{1}{2} log\frac{det[\Sigma_{\alpha_u}]}{det[\Sigma_1]} - \frac{D}{2}  + \frac{1}{2}  tr[\Sigma_{\alpha_u}^{-1}(\Sigma_1 + \mu_1\mu_1^T - 2\mu_{\alpha_u}\mu_1^T+\mu_{\alpha_u}\mu_{\alpha_u}^T)]  \\
&= \frac{1}{2} log\frac{det[\Sigma_{\alpha_u}]}{det[\Sigma_1]} - \frac{D}{2}  + \frac{1}{2}  tr[\Sigma_{\alpha_u}^{-1}\Sigma_1] + \frac{1}{2} tr[\mu_1^T\Sigma_{\alpha_u}^{-1}\mu_1 - 2\mu_1^T\Sigma_{\alpha_u}^{-1}\mu_{\alpha_u}+\mu_{\alpha_u}^T\Sigma_{\alpha_u}^{-1}\mu_{\alpha_u})]  \\
&=  \frac{1}{2} log\frac{det[\Sigma_{\alpha_u}]}{det[\Sigma_1]} - \frac{D}{2}  + \frac{1}{2}  tr[\Sigma_{\alpha_u}^{-1}\Sigma_1] + \frac{1}{2} (\mu_{\alpha_u} - \mu_1)^T\Sigma_{\alpha_u}^{-1}(\mu_{\alpha_u} - \mu_1).
\end{aligned}
\end{equation}

The calculation of $\mathbb{E}_{\mathcal{N}_2}[log{\mathcal{N}_2} - log{\mathcal{N}_{\alpha_u}}]$ is the same, and then, $JS^{G_{\alpha_u}}$ is given by:
\begin{equation}
\footnotesize		 
\begin{aligned}
JS^{G_{\alpha_u}} 
&= \frac{1-{\alpha_u}}{2} log\frac{det[\Sigma_{\alpha_u}]}{det[\Sigma_1]} - \frac{D(1-{\alpha_u})}{2}  + \frac{1-{\alpha_u}}{2}  tr[\Sigma_{\alpha_u}^{-1}\Sigma_1] + \frac{1-\alpha_u}{2} (\mu_{\alpha_u} - \mu_1)^T\Sigma_{\alpha_u}^{-1}(\mu_{\alpha_u} - \mu_1) + \\
&\frac{\alpha_u}{2} log\frac{det[\Sigma_{\alpha_u}]}{det[\Sigma_2]} - \frac{D\alpha_u}{2}  + \frac{\alpha_u}{2}  tr[\Sigma_{\alpha_u}^{-1}\Sigma_2] + \frac{\alpha_u}{2} (\mu_{\alpha_u} - \mu_2)^T\Sigma_{\alpha_u}^{-1}(\mu_{\alpha_u} - \mu_2)   \\
&=\frac{1}{2}(log\frac{det[\Sigma_{\alpha_u}]^{1-\alpha_u}}{det[\Sigma_1]^{1-\alpha_u}} + log\frac{det[\Sigma_{\alpha_u}]^{\alpha_u}}{det[\Sigma_2]^{\alpha_u}}) - \frac{D}{2} + \frac{1}{2}tr(\Sigma^{-1}_{\alpha_u}((1-\alpha_u)\Sigma_1+\alpha_u\Sigma_2)) + \\
& \frac{1-\alpha_u}{2} (\mu_{\alpha_u} - \mu_1)^T\Sigma_{\alpha_u}^{-1}(\mu_{\alpha_u} - \mu_1) +  \frac{\alpha_u}{2} (\mu_{\alpha_u} - \mu_2)^T\Sigma_{\alpha_u}^{-1}(\mu_{\alpha_u} - \mu_2) \\
& = \frac{1}{2}(log[\frac{det[\Sigma_{\alpha_u}]}{det[\Sigma_1]^{1-\alpha_u} det[\Sigma_2]^{\alpha_u}}] - D + tr(\Sigma^{-1}_{\alpha_u}((1-\alpha_u)\Sigma_1+\alpha_u\Sigma_2))+ (1-\alpha_u) (\mu_{\alpha_u} - \mu_1)^T\Sigma_{\alpha_u}^{-1}(\mu_{\alpha_u} - \mu_1) + \\
&  \alpha_u (\mu_{\alpha_u} - \mu_2)^T\Sigma_{\alpha_u}^{-1}(\mu_{\alpha_u} - \mu_2)).
\end{aligned}
\end{equation}

As to the dual form $JS_*^{G_{\alpha_u}}$, we calculate $\mathbb{E}_{\mathcal{N}_{\alpha_u}}[log{\mathcal{N}_{\alpha_u}} - log{\mathcal{N}_1}]$, which is given by:
\begin{equation}
\footnotesize
\begin{aligned}
\frac{1}{2} log\frac{det[\Sigma_1]}{det[\Sigma_{\alpha_u}]} - \frac{D}{2}  + \frac{1}{2}  tr[\Sigma_1^{-1}\Sigma_{\alpha_u}] + \frac{1}{2} ( \mu_1-\mu_{\alpha_u})^T\Sigma_1^{-1}(\mu_1-\mu_{\alpha_u}).
\end{aligned}
\end{equation}
Then, the calculation of $\mathbb{E}_{\mathcal{N}_{\alpha_u}}[log{\mathcal{N}_{\alpha_u}} - log{\mathcal{N}_2}]$ is the same, and $JS_*^{G_{\alpha_u}}$ is given by:
\begin{equation}
\footnotesize		 
\begin{aligned}
JS_*^{G_{\alpha_u}} 
&= \frac{1-\alpha_u}{2} log\frac{det[\Sigma_1]}{det[\Sigma_{\alpha_u}]} - \frac{D(1-\alpha_u)}{2}  + \frac{1-\alpha_u}{2}  tr[\Sigma_1^{-1}\Sigma_{\alpha_u}] + \frac{1-\alpha_u}{2} ( \mu_1-\mu_{\alpha_u})^T\Sigma_1^{-1}(\mu_1-\mu_{\alpha_u}) + \\
& \frac{\alpha_u}{2} log\frac{det[\Sigma_2]}{det[\Sigma_{\alpha_u}]} - \frac{D\alpha_u}{2}  + \frac{\alpha_u}{2}  tr[\Sigma_2^{-1}\Sigma_{\alpha_u}] + \frac{\alpha_u}{2} ( \mu_2-\mu_{\alpha_u})^T\Sigma_2^{-1}(\mu_2-\mu_{\alpha_u}) \\
& = \frac{1}{2} (log\frac{det[\Sigma_1]^{1-\alpha_u}}{det[\Sigma_{\alpha_u}]^{1-\alpha_u}} + log\frac{det[\Sigma_2]^{\alpha_u}}{det[\Sigma_{\alpha_u}]^{\alpha_u}}) - \frac{D}{2} + \frac{1}{2}tr((1-\alpha_u)\Sigma_1^{-1}\Sigma_{\alpha_u}+\alpha_u\Sigma_2^{-1}\Sigma_{\alpha_u})+\frac{1-\alpha_u}{2}\mu_1^T\Sigma_1^{-1}\mu_1 - \\
& \frac{1-\alpha_u}{2}\mu_1^T\Sigma_1^{-1}\mu_{\alpha_u} - \frac{1-\alpha_u}{2}\mu_{\alpha_u}^T\Sigma_1^{-1}\mu_1 +
\frac{1-\alpha_u}{2}\mu_{\alpha_u}^T\Sigma_1^{-1}\mu_{\alpha_u}
+\frac{\alpha_u}{2}\mu_2^T\Sigma_2^{-1}\mu_2 - 
 \frac{\alpha_u}{2}\mu_2^T\Sigma_2^{-1}\mu_{\alpha_u} \\
& - \frac{\alpha_u}{2}\mu_{\alpha_u}^T\Sigma_2^{-1}\mu_2 + \frac{\alpha_u}{2}\mu_{\alpha_u}^T\Sigma_2^{-1}\mu_{\alpha_u} \\
& = \frac{1}{2}log\frac{det[\Sigma_1]^{1-\alpha_u}det[\Sigma_2]^{\alpha_u}}{det[\Sigma_{\alpha_u}]} - \frac{D}{2} + \frac{1}{2}tr(\underbrace{((1-\alpha_u)\Sigma_1^{-1}+\alpha_u\Sigma_2^{-1})}_{\Sigma^{-1}_{\alpha_u}}\Sigma_{\alpha_u}) + \frac{1-\alpha_u}{2}\mu_1^T\Sigma_1^{-1}\mu_1 - \\
& (1-\alpha_u)\mu_1^T\Sigma_1^{-1}\mu_{\alpha_u} + \frac{1-\alpha_u}{2}\mu_{\alpha_u}^T\Sigma_1^{-1}\mu_{\alpha_u} + \frac{\alpha_u}{2}\mu_2^T\Sigma_2^{-1}\mu_2 - 
 \alpha_u\mu_2^T\Sigma_2^{-1}\mu_{\alpha_u} + \frac{\alpha_u}{2}\mu_{\alpha_u}^T\Sigma_2^{-1}\mu_{\alpha_u} \\
& = \frac{1}{2}log\frac{det[\Sigma_1]^{1-\alpha_u}det[\Sigma_2]^{\alpha_u}}{det[\Sigma_{\alpha_u}]} + \frac{1-\alpha_u}{2}\mu_1^T\Sigma_1^{-1}\mu_1 + \frac{\alpha_u}{2}\mu_2^T\Sigma_2^{-1}\mu_2 - \underbrace{((1-\alpha_u)\mu_1^T\Sigma_1^{-1} + \alpha_u\mu_2^T\Sigma_2^{-1})}_{\mu^T_{\alpha_u}\Sigma_{\alpha_u}^{-1}}\mu_{\alpha_u} + \\
&\frac{1}{2}\mu_{\alpha_u}^T\underbrace{((1-\alpha_u)\Sigma^{-1}_1+\alpha_u\Sigma^{-1}_2)}_{\Sigma^{-1}_{\alpha_u}}\mu_{\alpha_u} \\
& = \frac{1}{2}(log\frac{det[\Sigma_1]^{1-\alpha_u}det[\Sigma_2]^{\alpha_u}}{det[\Sigma_{\alpha_u}]} + (1-\alpha_u)\mu_1^T\Sigma_1^{-1}\mu_1 + \alpha_u\mu_2^T\Sigma_2^{-1}\mu_2 - {\mu^T_{\alpha_u}\Sigma_{\alpha_u}^{-1}}\mu_{\alpha_u}).
\end{aligned}
\end{equation}

\hfill $\square$

\section{Implementation Details}
The deep neural network configuration and training details are summarized in Table~\ref{tab:setting}. 
As for the NP-Match related hyperparameters, we set the lengths of both memory banks ($\mathcal{Q}$) to 2560. The coefficient ($\beta$) is set to 0.01, and we sample $T=10$ latent vectors for each target point. The uncertainty threshold ($\tau_u$) is set to 0.4 for CIFAR-10, CIFAR-100, and  STL-10, and it is set to 1.2 for ImageNet. NP-Match is trained by using stochastic gradient descent (SGD) with a momentum of 0.9.
The initial learning rate is set to 0.03 for CIFAR-10, CIFAR-100, and STL-10, and it is set to 0.05 for ImageNet.
The learning rate is decayed with a cosine decay
schedule \cite{loshchilov2016sgdr}, and NP-Match is trained for $2^{20}$ iterations. The MLPs used in the NP model all have two layers with $\mathcal{M}$ hidden units for each layer. For WRN, $\mathcal{M}$ is a quarter of the channel dimension of the last convolutional layer, and as for ResNet-50, $\mathcal{M}$ is equal to 256. To compete with the most recent SOTA method \cite{zhang2021flexmatch}, we followed this work to use the Curriculum Pseudo Labeling (CPL) strategy in our method and UPS \cite{rizve2021defense}. We initialize each memory bank with a random vector.  

\begin{table}[h]
\centering 
\resizebox{0.6\textwidth}{!}{
\begin{tabular}{@{}c|c|c|c|c@{}}
 \toprule[1pt]
Dataset  &  CIFAR-10  & CIFAR-100  & STL-10 & ImageNet \\
 \hline  
 Model &  WRN-28-2   & WRN-28-8  & WRN-37-2  & ResNet-50  \\
\hline
 Weight Decay &  5e-4 & 1e-3  & 5e-4   & 1e-4  \\
\hline
 Batch Size (B) & \multicolumn{3}{c|}{64} & 256 \\
 \hline
 $\mu$ &  \multicolumn{3}{c|}{7} &  1 \\
 \hline
 Confidence Threshold ($\tau_c$) &  \multicolumn{3}{c|}{0.95} &  0.7  \\
  \hline
 EMA Momentum&  \multicolumn{4}{c}{0.999}  \\
  \hline
 $\lambda_u$ &  \multicolumn{4}{c}{1.0}  \\
  \bottomrule[1pt]
  \end{tabular}}
 \caption{Details of the training setting.} 
 \label{tab:setting} 
 \end{table}

We ran each label amount setting for three times using different random seeds to obtain the error bars on CIFAR-10, CIFAR-100, and STL-10, but on ImageNet, we only ran for once. GeForce GTX 1080 Ti GPUs were used for the experiments on CIFAR-10, CIFAR-100, and STL-10, while Tesla V100 SXM2 GPUs were used for the experiments on ImageNet. 

\begin{figure}[t]
\centering
\includegraphics[width=\linewidth]{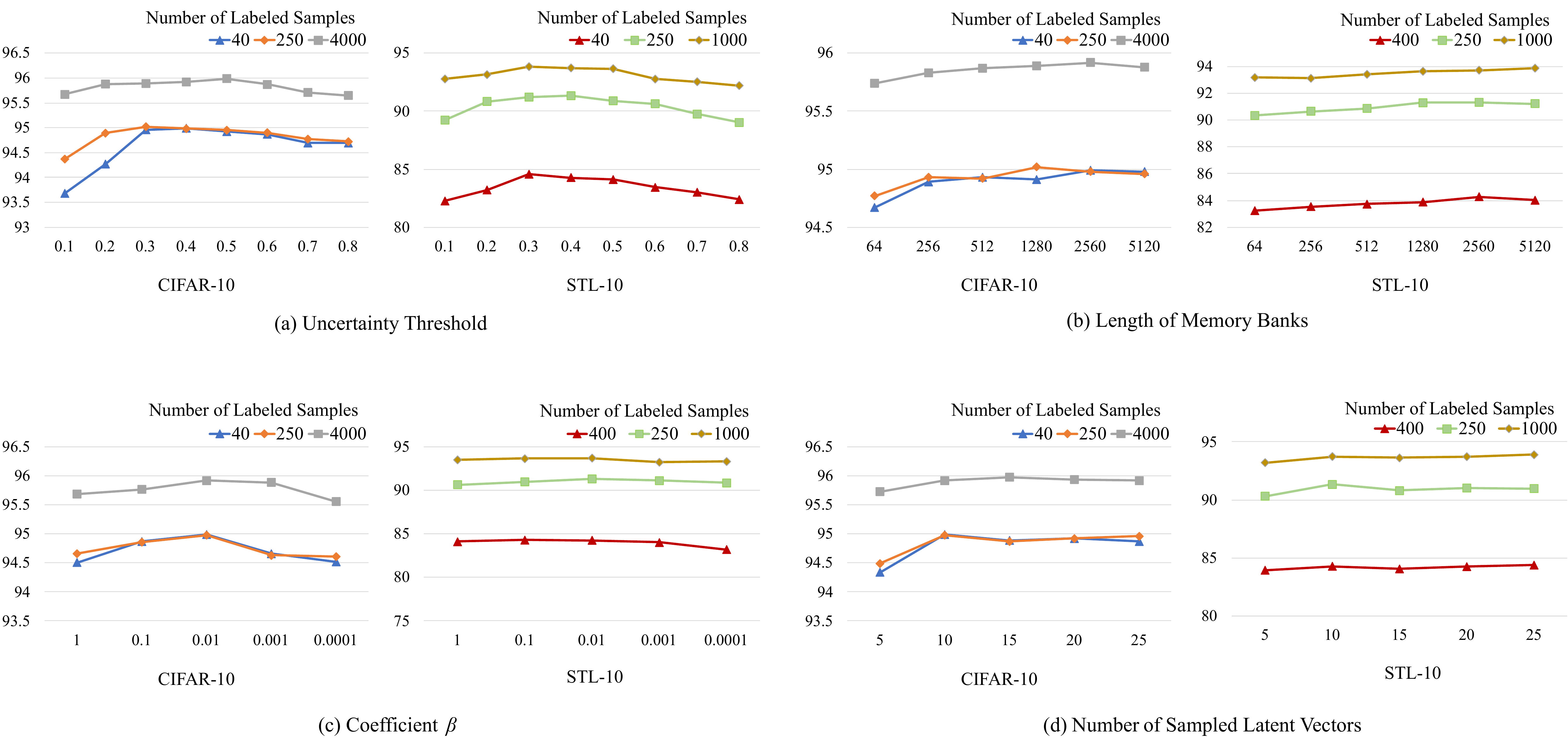}
\vspace{-4ex}
\caption{Performance for different hyperparameters.} 
\label{fig:hyper}
\end{figure}

\section{Hyperparameter Exploration}
We did additional experiments on CIFAR-10 and STL-10 in terms of the hyperparameters related to the NP model in NP-Match, in order to explore how performance is affected by the changes of hyperparameters, which may provide some hints to readers for applying NP-Match to other datasets. We consider four hyperparameters in total, including the uncertainty threshold ($\tau_u$), the length of memory banks ($\mathcal{Q}$), the coefficient of $JS^{G_{\alpha_u}}$ ($\beta$), and the number of sampled latent vectors ($T$). By Figure~\ref{fig:hyper}(a), a reasonable $\tau_u$ is important. Specifically, lower $\tau_u$ usually leads to worse performance, because lower $\tau_u$ enforces NP-Match to select a limited number of unlabeled data during training, which is equivalent to training the whole framework with a small dataset. Conversely, when $\tau_u$ is too large, more uncertain unlabeled samples are chosen, whose pseudo-labels might be incorrect, and using these uncertain samples to train the framework can also lead to a poor performance. Furthermore, the difficulty of a training set also affects the setting of $\tau_u$, as a more difficult dataset usually has more classes and hard samples (e.g., ImageNet), which makes the uncertainties of predictions large, so that $\tau_u$ should be adjusted accordingly. 
From Figure~\ref{fig:hyper}(b), the performance becomes better with the increase of $\mathcal{Q}$. When more context points are used, the more information is involved for inference, and then the NP model can better estimate the global latent variable and make predictions. This observation is consistent with the experimental results where the original NPs are used for image completion \cite{garnelo2018neural}.  
Figure~\ref{fig:hyper}(c) shows the ablation study of $\beta$ that controls the contribution of $JS^{G_{\alpha_u}}$ to the total loss function, and when $\beta=0.01$, we can obtain the best accuracy on both datasets. By  Figure~\ref{fig:hyper}(d), if $T$ is smaller than 5, then the performance will go down, but when $T$ is further increased, then the performance of NP-Match is not influenced greatly.

\end{document}